\documentclass[runningheads]{llncs}
\usepackage{graphicx}
\usepackage{hyperref}
\usepackage{multirow}
\usepackage{xcolor}
\usepackage{array}
\usepackage{caption}
\usepackage[linesnumbered,ruled,vlined]{algorithm2e} 
\usepackage{comment}

\begin{document}
\raggedbottom
%
%
\title{Interpretability of Blackbox ML Models through Dataview Extraction and Shadow Model creation}
%
\author{Rupam Patir \and Shubham Singhal\and C. Anantaram \and Vikram Goyal}
%
%
\institute{Indraprastha Institute of Information Technology Delhi 
\email{\{rupam13081,shubham18016,c.anantaram,vikram\}@iiitd.ac.in}\\
}
\titlerunning{Interpretability through Dataview Extraction and Shadow Models}
\maketitle              
\begin{abstract}
Deep learning models trained using massive amounts of data, tend to capture one view of the data and its associated mapping.
Different deep learning models built on the
same training data may capture different views of the data based on the underlying techniques used. For
explaining the decisions arrived by blackbox deep learning models, we argue that it is
essential to reproduce that model’s view of the training data faithfully. This faithful
reproduction can then be used for explanation generation. We investigate two methods for
data view extraction: hill-climbing approach and a GAN-driven approach. We then use this synthesized data for creating shadow models for explanation generation: Decision-Tree model and Formal Concept Analysis based model. We evaluate these approaches on a Blackbox model trained on public datasets
and show its usefulness in explanation generation.

\end{abstract}
\section{Introduction}

\indent With its use to make critical decisions in a wide range of industries from health to business, Machine Learning (ML) models are becoming more useful and significant by the day. Inevitably, the interpretability of these models has become a major focus as it is important to understand how the model arrives at its decisions. With sophisticated models such as Convolution Neural Networks (CNN) and Deep Neural Networks (DNN), the problem of interpretability becomes especially relevant when the user of the system is interested not only in the final decision but also in how and why it came to that decision. It is envisaged that interpretability is a promising approach for addressing challenges such as user trust issues due to algorithm aversion, data quality issues related to fairness and, to understand the reasoning behind decisions.

\indent An example of such a challenge is in the health care sector where it is crucial to understand why an algorithm predicted a disease with respect to an individual patient. Not only does this ensure that the user using the system has a modicum of confidence that the system is looking at the right aspects of the patients record when coming to its decision, but it also gives the user a chance to identify new indicators for a disease. Another example is in the business sector where a marketeer would like to predict repeat buyers for a product. Knowing how a system came to its conclusion not only increases the marketeer's confidence in the system but also gives valuable insight for critical business decisions.

\indent Suresh and Guttag \cite{ref14} have shown that building a ML system, \textbf{M}, can be viewed as a series of data transformations, wherein $X$ and $Y$ are the real-world underlying features and labels respectively that \textbf{M} should ideally model. However, since the real-world data may be huge and may not always be completely available, a large sample $X_{n}$ and $Y_{n}$ is taken to train the model \textbf{M}. During the construction of \textbf{M} itself,  inevitably the features $X_{n}$ and labels $Y_{n}$ are projected to some $X'_{n}$ and $Y'_{n}$ respectively in the model due to the way the model parameters, representation of the features and labels are chosen and trained. Thus instead of the original function f: $X \rightarrow Y$ , what is learned is the function g: $X'_{n} \rightarrow Y'_{n}$ . This $X'_{n}$ and $Y'_{n}$, we call as the \textit{data view} captured by the model.

In our paper, we focus on blackbox ML models wherein the model's details such as its parameters, its mapping and representation details are not available. In such a scenario, we propose that it is important to faithfully extract the  `data view' captured in the model \textbf{M} in order to be able to build interpretable models for that blackbox ML model \textbf{M}.  We define a notion of \textit{data view} of the target blackbox model, using the set of data objects correctly classified by the model. Our method extracts the data view via data synthesis to create a close approximation of the target model's \textit{data view}. 

We propose two different techniques for this step. The first technique, inspired by Hill-climbing, is a query synthesis technique that generates a dataset such that the output probability vectors has the least entropy as per the target model. The second technique, inspired by GAN, learns a model for the data generation such that the target labels classified using the blackbox model will have high accuracy. This is especially useful for synthesizing data for high dimensional datasets where the query synthesis technique becomes computationally expensive.

\indent Once the data view is extracted, our approach aspires to create an interpretable Shadow model on that data view based on which the Blackbox interpretability is achieved. We build Decision Tree model based on this Data View. We also use Formal Concept Analysis techniques to dive deeper into the interpretation of the target model. These shadow models then provide the interpretable models for that blackbox model.

Our main contributions in this paper are as follows: \begin{itemize} \item We show that the `data view' extracted from a Blackbox model is a better reflection of the model's behaviour, and that `data view' can be synthesized from the model.
\item We study two synthesis methods for data view extraction -- a Hill-climbing approach, and a new approach based on Generative Adversarial Networks (GAN).
\item Using this synthesized dataset, we can train interpretable models to generate explanations and interpret the original target model. Two interpretable models that we focus on are creating Decision Trees and interpretation via Formal Concept Analysis.

\end{itemize}

\section{State-of-the-art}

\indent Various approaches for interpretability of blackbox models have been proposed \cite{ref18,ref16,ref17,ref19}. Broadly, work on explainability  can be classified into three types: a) Model-inspection methods, b) Shadow-model methods, and c) Data-based methods.

\vspace{-0.4cm}

\subsubsection{Model-inspection methods} Class Activation Maps (CAM) and Gradient-based Class Activation Maps (GradCAM) \cite{ref20} inspect the deep network and compute a feature-importance map by associating the feature
maps in the final convolutional layer with particular classes.
In GradCAM the correlation of the gradients of each class w.r.t. each feature
map is done by weights of activations of the feature
maps as an indication of which inputs are most important.

\vspace{-0.4cm}

\subsubsection{Shadow-model methods}LIME \cite{ref1} relies on building a locally linear shadow model for interpretability. There are broadly two types of interpretability approaches - Local and Global. Local interpretability involves building shadow models that reflect model’s view on a localised input space. Ribeiro et al \cite{ref1} method called LIME (Locally Interpretable Model-Agnostic Explanations) assume that the complex learned function can be approximated by a set of locally linear models. Global interpretablity works on the entire input space, giving the target model’s global view on the data. Bastani et al \cite{ref2}, propose a shadow model approach but on a global scale wherein they approximate the target model in the form of an interpretable shadow model that is generated after fitting a Gaussian distribution to the training data.

\vspace{-0.4cm}

\subsubsection{Data-based methods}Lakkaraju et al \cite{ref13} explain the global view using Decision Sets. Sangroya et.al \cite{ref12} have used a Formal Concept Analysis based method to provide data based explanability.

In the above approaches either the availability of training or validation data is essential or the details of the Target model is essential. However, in blackbox models the assumption of availability of training data or having the details of the model cannot be assumed. Thus our approach differs in that we extract the data view captured in the model and then build interpretable shadow models for the blackbox models.

\vspace{-0.2cm}

\section{Methodology}

\vspace{-0.2cm}

\indent Our method takes a machine learning model \textbf{M} for which we have oracle access i.e. black box access where we have access to the input space \textbf{f} and the output vector of probabilities \textbf{p}. However, no information is available regarding the original data distribution and model parameters. We call the model \textbf{M} as the target model. To explain the target model we create an interpretable shadow model that simulates the functionality of the target model in terms of decision making over input data instances. We use the following steps for the same.


\begin{enumerate}
    \item Synthesize a data view to be used for training a shadow model.
    \item Train an interpretable shadow model on the synthesized data view.
    \item Transform the model into a set of rules. 
\end{enumerate}

\subsection{Data View generation through data synthesis}
As discussed earlier, a blackbox ML model \textbf{M} captures a particular `Data View' of the original training data, and models the mapping from input features to labels/ predictions through some complex non-linear function. For a multi-class classification problem, this is viewed as classifying clearly positive instances in its appropriate class with some instances falling in the boundary regions between the classes. If we take only the clearly positive instances being classified in each of the classes with a confidence threshold of 0.7 and above and leave out all instances that are very close to the borders of the classes with less confidence, then a simpler interpretable shadow model can be constructed for the blackbox model. We argue that it is better to synthesize data for the ML model \textbf{M}, than to take the original training data, since the ML model \textbf{M} has captured a `data view' from the original data. Thus it is more appropriate to extract out that data view and use it to train an interpretable shadow model \textbf{S}. We do this `data synthesis' by generating data instances and posing these instances to \textbf{M} for classification / prediction, and considering only those instances that have a positive classification / prediction beyond a threshold of 70\% by the model. The generated dataset then reproduces the data view of the model \textbf{M}.

\indent We explore two different techniques for the data synthesis. The first technique generates a dataset whose output probability vectors has the least entropy as per the target model \textbf{M}. On the other hand the second technique learns a model for the data generation.

\vspace{-0.4cm}

\subsubsection{Synthesis using Hill Climbing Method.}
\indent The first technique uses a hill climbing algorithm to synthesize data \textbf{D}, which we will use to train our interpretable model \textbf{S}.

To build \textbf{D} we use a query synthesis algorithm as proposed in the Membership attack paper by Shokri et al \cite{ref4}. Here, we use a function Synthesize-Record to generate one record for a class label \textbf{c}. It generates a record that assigns random values for each feature in the model’s input space. We then feed this record to the model \textbf{M} and get it’s class probability vector \textbf{p}. The algorithm accepts a record as part of our dataset \textbf{D} only if our model is confident beyond a threshold (conf$_{min}$) that the record belongs to a class \textbf{c}. If the record does not meet this threshold the algorithm randomly reassigns  \textbf{k} features and repeat the process. Each time the record gets rejected, the algorithm reduces the value of \textbf{k} and repeat the process. If there are no more features left to be re-assigned, the algorithm discards the record and start again from the beginning. We repeat this process such that we have a significant amount of records per class label to train our interpretable model \textbf{S}. The algorithm also limits the number of features it reassigns with k$_{max}$ and k$_{min}$.  The algorithm uses these limitations to speed up the process and considers those records which will actually be admitted to the database. The algorithm reduces the number of features \textbf{k} in each revision so that the permutations are localised to the record being considered. The randomize function is specific to every dataset, where we randomize with respect to whatever knowledge we have of the input space.


\subsubsection{Synthesis with GAN.}
The Hill Climbing Method would be costly when synthesizing highly dimensional datasets. We propose a method based on GAN to generate synthetic data which can be used for both low and high dimensional datasets. We first design a neural net architecture to be used as a generator for each of the dataset mentioned in table \ref{tab:dataset}.
The goal of this generator is to generate a data point given to any corresponding noise. The number of input and output neurons of the generator is $noise\_size$ and number of features in the dataset respectively.

\begin{figure*}[h]
\vspace{-1cm}
\includegraphics[width=\textwidth]{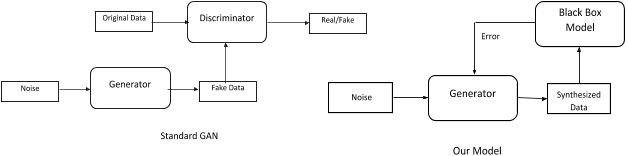}
\caption{Flow Chart of the GAN Approach.} \label{fig3}

\end{figure*}

The steps for Synthesizing the data are as follows:
\begin{enumerate}
    \item For each class \textbf{c} we make a different generator \textbf{$G_{c}$} that takes a random noise of size $noise\_size$(depends on the dataset) as an input and generates a data sample as an output. Steps 2-4, as shown in algorithm \ref{alg:GAN}, is repeated for each generator \textbf{$G_{c}$}.
    \item For training the generator we replace what would normally be an input image and discriminator in a standard GAN model with our Black box model \textbf{M} as shown in the figure \ref{fig3}.
    \item In the Forward pass, we generate random noise and taking that as input we generate a synthesized record and then pass it to the Black box model \textbf{M} to get how close it is to the real record.
    \item In the Backward pass, first freeze the weights \textbf{w$_{bbm}$} of the Black box model \textbf{M} and then propagate the error $\epsilon$ generated at output of Black box model \textbf{M}. Using the error obtained at input layer of Black box model \textbf{M}, we update the weights \textbf{w$_{G_{c}}$} of the Generator \textbf{$G_{c}$} by making it the error obtained at the output layer of the generator \textbf{$G_{c}$} and back propagating it.
    \item After training the generator \textbf{$G_{c}$} we can generate data by inputting any random noise of size $noise\_size$ and getting the output.
    
\end{enumerate}

\vspace{-0.4cm}

\begin{algorithm}[h]
\SetAlgoLined

\KwResult{Trained Generator for class c}
\SetKwFunction{FMain}{Synthesize}
\SetKwProg{Fn}{Function}{:}{}
\caption{Training Generator for generating data of class c}
\label{alg:GAN}
\DontPrintSemicolon
\Fn{\FMain{$BlackBoxModel$,$c$,$epochs$,$batch\_size$,$noise\_size$,$num\_class$}}{
    $generator \leftarrow createGenerator()$\;
    $gan \leftarrow Combine(generator,BlackBoxModel)$\;
    $FreezeWeights(BlackBox)$\;
    \For{$(iteration = 0; iteration < epochs; iteration++)$}{
        $noise \leftarrow random(shape \leftarrow (batch\_size,noise\_size))$\;
        $out \leftarrow zeros(batch\_size,num\_class)$\;
        $out[c] \leftarrow 0.99$\;
        $gan.train(noise,out)$\;
     }
    $Return$ $generator$
  }

\end{algorithm}

\subsection{Shadow Model and its Fidelity} 
Now that we have our synthesized dataset, we use it to train our Interpretable model \textbf{S} which can be a Decision tree or any other easy-to-understand model.\\
To check the fidelity (similarity) of our shadow model i.e. our Interpretable model, we calculate the number of records that are predicted the same by both the Black Box model \textbf{M} and the Interpretable model \textbf{S}.

\begin{equation}
Fidelity = \sum_i \frac{C_{i}}{n}
\end{equation}

where, $C_{i}$ = 0 if $Classification(S,i) \neq Classification(M,i)$, and $C_{i}$ = 1 if $Classification(S,i) = Classification(M,i)$ and \textit{n} = total number of records.

\subsection{Interpretability}
\indent If our shadow model and target model have a high level of fidelity we can be confident enough to assert that the easy-to-interpret shadow model is a close approximation of our hard-to-interpret target model. As such, with our shadow model we can now run commonly known interpretation techniques such as viewing it as a decision tree. We can also get an understanding of feature importance within the model. For example, with Decision trees, we can interpret the importance of features depending on how high up the tree the feature causes a split. Therefore, depending on the choice of the shadow model, we can run different methods to better interpret our target model thereby achieving our goal.\\
\indent For feature importance we can use the method of Permutation Importance. Permutation Importance is a method of finding the important features in a classification model. It involves permuting the data to see which features have the largest effect on the accuracy of our model. For example, if permuting feature $A$ decreases our accuracy or results in a change of classification, we can imply that feature $A$ is important in the classification process of the class label of that record.\\
\indent We also use Formal Concept Analysis (FCA) using techniques proposed by Sangroya et al \cite{ref12} to carry our different analytical procedures. Specifically, we use FCA to produce implication rules and as another metric to evaluate the feature importance in the target model's view.

\vspace{-0.3cm}

\section{Experimental Results}

\vspace{-0.2cm}

We consider Neural Networks as target models and learn them using Python's Keras library for each of the datasets given in Table \ref{tab:dataset}. For GAN approach the target model is trained on scaled data with values between -1 to 1. 
Our goal is to investigate the effect of using a data view of a target model on the fidelity of Shadow model. Does the use of the data-view of the target model result into better fidelity of a shadow model? To demonstrate this, we create two shadow models; first shadow model, $OShadow$, uses the original data that was used for training the target model, the second model, $SShadow$, uses the data view of the target model for its training.  We demonstrate through experiments that the $SShadow$ model outperforms $OShadow$ consistently in terms of fidelity. The experimental results on fidelity and accuracy are given in Table \ref{tab1:results}. The table also shows the accuracy of $SShadow$ model on the test data to get an idea of how much the shadow model matches with the target model in terms of generalizability. We also study the effect of different parameters on the fidelity of a shadow model for the Purchase dataset, which is explained further.

\newcolumntype{D}[1]{>{\centering\arraybackslash}m{#1}}

 \begin{minipage}{\textwidth}
  \begin{minipage}[b]{0.35\textwidth}
    \vspace{1.5cm}
    \includegraphics[width=1.\linewidth,height=3.5cm]{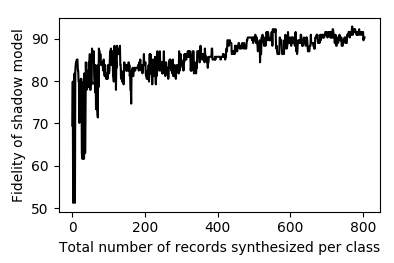}
    \captionof{figure}{Fidelity of SShadow vs Number of records synthesized} 
    \label{fig:fidvsrec}
  \end{minipage}
    \hspace{0.1cm}
    \begin{minipage}[t]{0.6\textwidth}
        \vspace{-6cm}
        \centering
        \begin{tabular}{|D{1.5cm}|D{1.5cm}|D{2cm}|D{2.1cm}|}
        \hline
        Dataset Name & \#objects & \#attributes & \# Type of Prediction \\ \hline
        Animal\footnotemark[1] & 100  & 16 & 7 class classification\\\hline
        Diabetes\footnotemark[2] & 768 & 8 & Probabilty score for Diabetes \\ \hline
        Mobile\footnotemark[3] & 2000 & 20 & Price range \\\hline
        Income\footnotemark[4] & 32561 & 107 & 2 class classification \\ \hline
        Purchase\footnotemark[5] & 9600 & [10,20,40,75] & Labeled classification by KNN\\ \hline
        \end{tabular}
        \captionof{table}{Datasets used for expirments}.
        \label{tab:dataset}
    \end{minipage}
  \end{minipage}
  
\begin{tiny}
\footnotetext[1]{\url{https://archive.ics.uci.edu/ml/datasets/Zoo}}
\footnotetext[2]{\url{https://www.kaggle.com/uciml/pima-indians-diabetes-database}}
\footnotetext[3]{\url{https://www.kaggle.com/iabhishekofficial/mobile-price-classification}}
\footnotetext[4]{\url{https://www.kaggle.com/uciml/adult-census-income}}
\footnotetext[5]{\url{https://www.kaggle.com/c/acquire-valued-shoppers-challenge/data}}
\end{tiny}

\subsection{Datasets and Target Model Learning}
\indent To analyse our algorithms we use the datasets described in Table \ref{tab:dataset}. 

Income dataset originally had 14 features which increase to 107 after binarization of the categorical features for training the target Neural Network. The reason to use the Purchase dataset to study the effect of aforementioned parameters lies in its flexibility to define the number of classes and dimensions. Purchase dataset is a user-product table which we cluster to get the labels. The number of clusters defines the number of class labels. Similarly it is easy to also choose number of products in the database randomly that defines the number of dimensions.

We synthesize 10000 records for each class label for each of our datasets using two approaches: Hill Climbing and GAN. A good data view should have mainly two properties: i) it should span the input domain of the application as much as possible and, ii) it should include the core objects of each class. We find that for the studied datasets, 10000 records satisfy both of these properties. Furthermore, from Figure \ref{fig:fidvsrec} we see that increasing the number of records does not negatively impact our fidelity. By default we use a Confidence Threshold of 70 percent for each of our synthesized datasets as this gives us records that the target model is fairly to very confident that the records belong to that class. Furthermore, high threshold values may limit the types of records to only those that model is very confident resulting in underfitting, whereas, low threshold values may permit records that have no valuable information about the model's view.

\subsection{Results}

We summarise our results of fidelity for all the datasets in Table \ref{tab1:results}. With the Hill Climbing algorithm, we see that the fidelity is higher when a shadow model is trained on the target view. The only exception to this is the result for Purchase (30F \& 2C) Dataset which in contrast to Purchase (20F \& 5C) has lower SShadow model fidelity. It may be so because of the relatively higher number of features which affects the fidelity of our SShadow model. We can see from Table \ref{tab1:results} that GAN algorithm does not work well for Mobile and Purchase datasets. This is due to the case that generator is not able to generate data with much varying confidence thus reducing the variance in the generated data. Also in purchase dataset the values of features is either 0 or 1 which leads to sparsity within data and thus generator is not able to learn the data view correctly. 

\newcolumntype{C}[1]{>{\centering\arraybackslash}m{#1}}
\begin{table}[h]
    \centering
    \caption{Evaluating the accuracy and fidelity of our target and Shadow models on different datasets. S stands model trained for scaled data used for GAN approach}
    \begin{tabular}{|C{2cm}|C{2cm}|C{2cm}|C{1.3cm}|C{1.3cm}|C{1.3cm}|C{1.3cm}|}
    \hline    
    \textbf{Dataset} & \textbf{Accuracy of Target Model} & \textbf{Fidelity OShadow} & \multicolumn{2}{|C{2.6cm}|}{\textbf{Fidelity SShadow})} & \multicolumn{2}{|C{2.6cm}|}{\textbf{Accuracy of SShadow}} \\
    \hline 
     & & & Hill & GAN & Hill & GAN\\
     \hline
     Animal & 95.23\newline 95.23(S) & 95.23\newline 95.23(S) & 95.23 & 95.23 & 90.47 & 90.47\\
     \hline
     Diabetes & 69.48\newline74.67(S) & 72.72\newline 80.51(S) & 84.41 &  81.81 & 71.42 & 74.67\\
     \hline
     Mobile & 94 & 84.75 & 92.25 & 62.74 & 90.25 & 64.25  \\
     \hline
     Income & 84.24 \newline 84.75(S) & 83.49 \newline 83.70(S) &  98.66 &  93.59 & 74.5 &  76.4 \\
     \hline
     Purchase (30F \& 2C) & 96.35 & 95.36 & \textbf{85.17} & 71.07 & 93.70 & 70.97 \\
     \hline
     Purchase (20F \& 5C) & 95.31 & 90.94 & 93.96 & 58.63 & \textbf{93.65} & 58.68 \\
    \hline
    \end{tabular}
    \label{tab1:results}
\end{table}

\newcolumntype{D}[1]{>{\centering\arraybackslash}m{#1}}
\begin{table}
\parbox{.3\linewidth}{
\centering
\caption{ Effect of the number of class labels}
\begin{tabular}{|D{1.2cm}|D{1.2cm}|D{1.2cm}|}
    \hline    
    \textbf{Classes} & \multicolumn{2}{|D{2.4cm}|}{\textbf{Fidelity}}\\
    \hline 
    & \textbf{Hill} & \textbf{GAN} \\
    \hline
    2 & 93.28 & 74.71\\
    5 & 99.01 & 59.625\\
    10 & 99.47 & 29.032\\
    15 & 99.73 & 27.99\\
    \hline
    \end{tabular}
    \label{tab2:classlabels}
}
\hspace{0.4cm}
\parbox{.3\linewidth}{
\centering
\caption{ Effect of the number of features}
\begin{tabular}{|D{1.4cm}|D{1cm}|D{1cm}|}
\hline    
\textbf{Features} & \multicolumn{2}{|D{2cm}|}{\textbf{Fidelity}}\\
\hline 
& \textbf{Hill} & \textbf{GAN} \\
\hline
10 & 100 & 100\\
20 & 85.01 & 72.94\\
30 & 81.11 & 71\\
40 & 75.91 & 73.08\\
50 & 67.58 & 72\\
75 & 52.13 & 71.59\\
\hline
\end{tabular}
\label{tab3:features}
}
\hspace{0.2cm}
\parbox{.3\linewidth}{
\centering
\caption{ Time taken for synthesis per record (in seconds)}
\begin{tabular}{|D{1.3cm}|D{1.1cm}|D{1.2cm}|}
\hline    
\textbf{Dataset} & \multicolumn{2}{|D{2.3cm}|}{\textbf{Time taken}}\\
\hline 
& \textbf{Hill} & \textbf{GAN} \\
\hline
Diabetes & 0.007 & 0.000075\\
Animal & 0.05 & 0.000127\\
Purchase (30F2C) & 0.02 & 0.00021\\
Mobile & 0.01 & 0.0001\\
Income & 0.03 & 0.00004\\
\hline
\end{tabular}
\label{tab3:timetaken}
}

\end{table}

\subsubsection{Effect of the number of Classes:}

To check the effect of varying the number of classes, we use the Purchase dataset with 15 features and generate 1000 records with 70\% confidence threshold. From Table \ref{tab2:classlabels}, we see that an increase in the number of classes does not negatively affect fidelity. This is so because the amount of information leaked by the target model would not decrease with an increase in the number of classes. Even when we have more classes, each feature retains and simultaneously leaks information about the decision boundaries of those classes. As the number of classes increases, the decision boundary gets more complex and generator is not able to generate data with high confidence.


\subsubsection{Effect of the number of Dimensions:}
To check the effect of varying the number of dimensions, we use the Purchase dataset with 2 classes and generate 1000 records with 70\% confidence threshold. From Table \ref{tab3:features}, we see that with the hill climbing method of data synthesis, increasing the dimensionality has a negative effect on fidelity. This is so because the synthesis process does not correctly capture the importance of features to the model’s view. For example, we may have more records varying in the unimportant features than in the important features giving us an unrepresentative view of the target model. On the other hand, there is not much effect on number of dimensions on GAN algorithm as our generator's architecture is also changing with respect to the number of dimensions. For more number of dimensions, we tend to have a more complex model.


\subsubsection{Effect of choice of synthesis process:}

Although the GAN process of synthesizing data produces better Synthesized fidelity than our hill climbing algorithm (From table \ref{tab1:results}), there is less coverage of the input space by the synthesized data. To induce better coverage within the synthesized data, we train multiple generators as each generator gives a different view captured by our target model. The data view we get is dependent upon how the weights of the generator are initialized. In case of hill climbing approach as we randomly initialize each data point generated, we get much more coverage within the data generated.

As seen from Table \ref{tab3:timetaken}, the time taken by the hill climbing approach is much more than the time taken by the GAN approach and also we can generate much more in using GAN approach as when a generator is fully trained we can get a new data point by inputting any random noise sample. 


\subsection{Interpreting our Shadow Models}


For using our shadow models to carry out explanations and visualisations we can use a varied number of techniques. To show that we can use a varied number of shadow models we use the Diabetes Dataset.


\subsubsection{Visualisation Using Decision Trees}

\begin{figure}[h]
\centering
\includegraphics[scale=0.24]{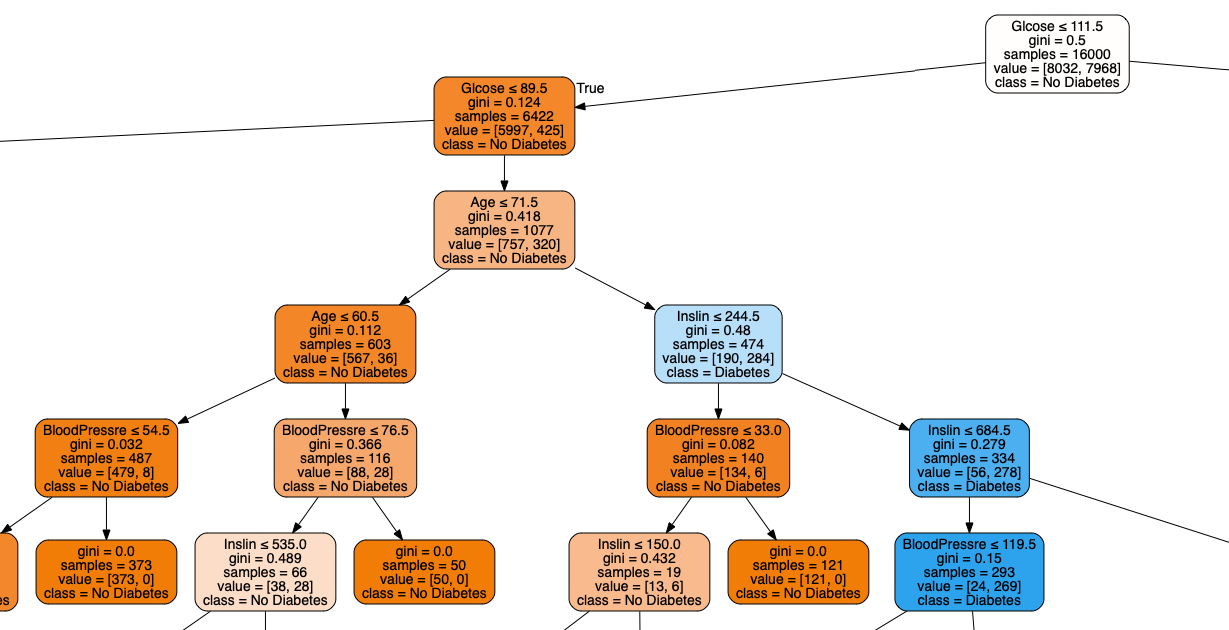}
\caption{Snippet of Shadow Model in the form Decision Tree trained on Synthesized Data generated from diabetes dataset} \label{fig:decisiontree}
\vspace{-0.6cm}
\end{figure}

With a target model trained on the diabetes dataset, we use our approach to train a shadow model in the form of a decision tree with which we can derive global or individual record decision rules. A snippet of the decision tree is shown in figure \ref{fig:decisiontree}. From our decision tree we see that Glucose is the root note and therefore is the most important feature.


\subsubsection{Permutation Importance}
We use the eli5\footnote[6]{\url{https://eli5.readthedocs.io/en/latest/}} library of Python to test Permutation Importance. On the Diabetes Dataset we found that the important features are the same in both our Target Model trained on the original data and our Shadow Model trained on the synthesized data. We find that the most important features are Glucose, BloodPressure, Insulin and Age.


\subsubsection{Formal Concept Analysis}
  For carrying out our analysis, we  binarize the features of our synthesized dataset as described in the paper by Sangroya et al \cite{ref12}. For example, a binary feature Insulin1 corresponds to an insulin level below 16 mIU. We use the Concepts.py\footnote[7]{\url{https://pypi.org/project/concepts/}} library in python to create formal concepts and the corresponding lattices for each class label using this data. From these lattices we carry out predictions based on the intents of the class lattices and intents of the individual records in our synthesized dataset. We present the feature importances in the model's view as calculated using the FCA approach in table \ref{tab7}. We also use ConExp\footnote[8]{\url{http://conexp.sourceforge.net/users/index.html}} to derive the implication rules that are considered during a classification, some of which are shown in table \ref{tab8}. 

\begin{table}
\parbox{.46\linewidth}{
    \caption{ Feature Importance using Formal Concept Analysis}
    \begin{tabular}{|D{1.6cm}|D{1.2cm}|D{1.2cm}|D{1.2cm}|}
    \hline    
    Feature & Diabetes & No Diabetes & Average\\
    \hline 
    Glucose & 92.19 & 35.29 & 63.74\\
    \hline    
    Blood Pressure & 78.13 & 27.21 & 52.67 \\
    \hline    
    Insulin & 81.25 & 21.32 & 51.285\\
    \hline    
    Age &  70.31 & 22.79 & 46.55\\
    \hline    
    Skin Thickness & 46.88 & 19.85 & 33.365 \\
    \hline    
    BMI & 40.63 & 20.59 & 30.61 \\
    \hline    
    Pregnancies &  32.81 & 22.79 & 27.8 \\
    \hline    
    DPF & 29.69 & 18.38 & 24.035 \\
    \hline
    \end{tabular}
    \label{tab7}
}
\hspace{0.5cm}
\parbox{.5\linewidth}{
    \centering
    \caption{ Implication rules derived via Formal Concept Analysis}
    \begin{tabular}{|D{3cm}|D{3cm}|}
    \hline    
    Rule & Translation\\
    \hline 
    Insulin2 $\rightarrow$ Class0 & An insulin level between 16 to 166 mIU L implies no diabetes\\
    \hline    
    Insulin3, Glucose2, Age3 $\rightarrow$ Class1 & Insulin level above \newline166  mIU L and Glucose level 140-200 mg dL and an age above 60 implies diabetes\\
    \hline    
    BP2, Age2 $\rightarrow$ Class0 & A blood pressure between 60 and 90 and an age between 20 and 60 implies no diabetes \\
    \hline
    \end{tabular}
    \label{tab8}
}
\hfill
\end{table}

\section{Analysis and Evaluation of Results}


From our shadow models, we found that in most cases whether we used Decision Trees, Random Forest Regressors or Formal Concept Analysis \cite{ref12}, the fidelity between the target model and the shadow model was better when trained with the synthesized data over the target model’s training data. This means that a shadow model is able to capture the view of a black box target model even without access to it's training data when using the target model itself to generate data to train the shadow model. However, certain datasets were relatively resistant to our synthesis processes such as datasets with a high number of features for the Hill Climbing approach and datasets with a high number of classes for the GAN approach. 

\indent With our shadow models, we also saw the different methods of interpretation that can be carried out on this shadow model and correspondingly the target black box model. From datasets with a low number of features, a Decision Tree approach can be used where we can see the affect of each feature in the classification process. The problem of interpretability becomes harder when considering datasets with large number of features. But given that we have a well approximated shadow model, other forms of interperation such as feature importance and implication rules can be used to provide interpretable explanations. Our approach can therefore be used with a multitude of choices for the intepretable shadow model and the method of interpretation can vary depending on the dataset. With our approach the view captured by the shadow model will tend to be more faithful to the target model's view and can therefore be used to explain and interpret the target model's view of the data.

\section{Conclusion}

From our experiments we have shown that via the procedure of synthesizing our dataset using the target black box model's predictions, we can create an approximation of the target model's view of the data to interpret and explain it's classification process. We also presented a synthesis process using a GAN that is useful for high dimensional data synthesis. We presented the different factors that affect the fidelity of our shadow models with each of our synthesis processes. Finally, we presented how we can use a variety of shadow model choices to interpret and understand our target model.

\bibliographystyle{splncs04}
\bibliography{rbib.bib}

\begin{thebibliography}{10}
\providecommand{\url}[1]{\texttt{#1}}
\providecommand{\urlprefix}{URL }
\providecommand{\doi}[1]{https://doi.org/#1}

\bibitem{ref18}
{Adadi}, A., {Berrada}, M.: Peeking inside the black-box: A survey on
  explainable artificial intelligence (xai). In: \textit{IEEE Access} (2018)

\bibitem{ref12}
Amit~Sangroya, C.~Anantaram, M.R., Rastogi, M.: Using formal concept analysis
  to explain black box deep learning classification models. In:
  \textit{IJCAI-19 Workshop, What can FCA do for Artificial Intelligence?}
  (2019)

\bibitem{ref7}
Awudu, K., Zhou, S.: {X-TREPAN:} a multi class regression and adapted
  extraction of comprehensible decision tree in artificial neural networks.
  CoRR  \textbf{abs/1508.07551} (2015)

\bibitem{ref2}
Bastani, O., Kim, C., Bastani, H.: Interpretability via model extraction. CoRR
  \textbf{abs/1706.09773} (2017)

\bibitem{ref10}
Chandrasekaran, V., Chaudhuri, K., Giacomelli, I., Jha, S., Yan, S.: Model
  extraction and active learning. CoRR  \textbf{abs/1811.02054} (2018)

\bibitem{ref16}
Chen, X., Duan, Y., Houthooft, R., Schulman, J., Sutskever, I., Abbeel, P.:
  Infogan: Interpretable representation learning by information maximizing
  generative adversarial nets. In: \textit{Advances in Neural Information
  Processing Systems (NIPS)} (2016)

\bibitem{ref5}
Craven, M.W., Shavlik, J.W.: Extracting tree-structured representations of
  trained networks. In: \textit{Proceedings of the 8th International Conference
  on Neural Information Processing Systems}. NIPS'95 (1995)

\bibitem{ref11}
Friedman, J.H., Popescu, B.E.: Predictive learning via rule ensembles. In:
  \textit{The Annals of Applied Statistics}. Institute of Mathematical
  Statistics (2008)

\bibitem{ref6}
Gilpin, L.H., Bau, D., Yuan, B.Z., Bajwa, A., Specter, M., Kagal, L.:
  Explaining explanations: An overview of interpretability of machine learning.
  In: \textit{2018 IEEE 5th International Conference on Data Science and
  Advanced Analytics (DSAA)} (2018)

\bibitem{ref17}
Guo, W., Huang, S., Tao, Y., Xing, X., Lin, L.: Explaining deep learning models
  -- a bayesian non-parametric approach. In: \textit{Proceedings of the 32Nd
  International Conference on Neural Information Processing Systems}. NIPS'18
  (2018)

\bibitem{ref13}
Lakkaraju, H., Bach, S.H., Leskovec, J.: Interpretable decision sets: A joint
  framework for description and prediction. In: \textit{Proceedings of the 22Nd
  ACM SIGKDD International Conference on Knowledge Discovery and Data Mining}.
  KDD '16 (2016)

\bibitem{ref15}
Murdoch, W.J., Singh, C., Kumbier, K., Abbasi-Asl, R., Yu, B.: Definitions,
  methods, and applications in interpretable machine learning. In:
  \textit{Proceedings of the National Academy of Sciences} (2019)

\bibitem{ref3}
Puri, N., Gupta, P., Agarwal, P., Verma, S., Krishnamurthy, B.: {MAGIX:} model
  agnostic globally interpretable explanations. CoRR  \textbf{abs/1706.07160}
  (2017)

\bibitem{ref1}
Ribeiro, M.T., Singh, S., Guestrin, C.: "why should {I} trust you?": Explaining
  the predictions of any classifier. In: \textit{SIGKDD} (2016)

\bibitem{ref20}
Selvaraju, R.R., Cogswell, M., Das, A., Vedantam, R., Parikh, D., Batra, D.:
  Grad-cam: Visual explanations from deep networks via gradient-based
  localization. In: \textit{IJCV} (2019)

\bibitem{ref4}
{Shokri}, R., {Stronati}, M., {Song}, C., {Shmatikov}, V.: Membership inference
  attacks against machine learning models. In: \textit{2017 IEEE Symposium on
  Security and Privacy (SP)} (2017)

\bibitem{ref19}
Shrikumar, A., Greenside, P., Kundaje, A.: Learning important features through
  propagating activation differences. CoRR  \textbf{abs/1704.02685} (2017)

\bibitem{ref14}
Suresh, H., Guttag, J.V.: A framework for understanding unintended consequences
  of machine learning. CoRR  \textbf{abs/1901.10002} (2019)

\bibitem{ref9}
Wang, F., Rudin, C.: Falling rule lists. In: \textit{AISTATS} (2015)

\end{thebibliography}
\nocite{*}

\end{document}